\documentclass[10pt,twocolumn,letterpaper]{article}

\usepackage{cvpr}              
\usepackage[accsupp]{axessibility} 
\usepackage{amsmath}
\usepackage{amssymb}
\usepackage{amsfonts,bm}
\usepackage{comment}

\usepackage[dvipsnames]{xcolor}

\definecolor{cvprblue}{rgb}{0.21,0.49,0.74}
\usepackage[pagebackref,breaklinks,colorlinks,citecolor=cvprblue]{hyperref}

\def\paperID{11} 
\def\confName{CVPR}
\def\confYear{2024}

\title{Adapting the Segment Anything Model During Usage in Novel Situations}

\author{Robin Schön \and Julian Lorenz \and Katja Ludwig \and Rainer Lienhart \\
Universität Augsburg\\
86159 Augsburg, Universitätsstraße 6a\\
{\{robin.schoen, julian.lorenz, katja.ludwig, rainer.lienhart\}@uni-a.de}
}

\begin{document} 
\maketitle
\begin{abstract}
 
The interactive segmentation task consists in the creation of object segmentation masks based on user interactions. The most common way to guide a model towards producing a correct segmentation consists in clicks on the object and background. The recently published Segment Anything Model (SAM) supports a generalized version of the interactive segmentation problem and has been trained on an object segmentation dataset which contains 1.1B masks. Though being trained extensively and with the explicit purpose of serving as a foundation model, we show significant limitations of SAM when being applied for interactive segmentation on novel domains or object types. On the used datasets, SAM displays a failure rate $\text{FR}_{30}@90$ of up to $72.6 \%$.
Since we still want such foundation models to be immediately applicable, we present a framework that can adapt SAM during immediate usage. For this we will leverage the user interactions and masks, which are constructed during the interactive segmentation process. We use this information to generate pseudo-labels, which we use to compute a loss function and optimize a part of the SAM model. The presented method causes a relative reduction of up to $48.1 \%$ in the $\text{FR}_{20}@85$ and $46.6 \%$ in the $\text{FR}_{30}@90$ metrics.

\end{abstract}    
\section{Introduction}
\label{sec:intro}
Many computer vision systems need object segmentation masks for single images as training material. The development of such  systems has especially been aided by the existence of large datasets for regular consumer images, such as COCO \citep{lin2014microsoft} and ADE20k \citep{zhou2017scene}. Some segmentation tasks, however, need much more specific data. Example domains for such cases are sports \citep{ludwig2023detecting, ludwig2023all}, agriculture \citep{roggiolani2023hierarchical},  medical image segmentation \citep{datasetkvasirseg}, and robotic vision \citep{zhuang2023instance}. 

The annotation of instance segmentation datasets usually incurs a high effort. Not only is there a large cost associated for human annotators, but in some difficult cases the creation of a high-quality mask is a non-negligible problem. An example for this would be the annotation of mask polygons when the object edges are finely jagged. In consequence, this led to the development of interactive segmentation systems. Such systems receive a simple, low-effort user interaction to create masks. This usually happens in an iteratively interactive context: The human refines computed masks by repeatedly interacting with the system, adding progressively more guiding interactions while inspecting the mask. This process goes on until the user is satisfied with the quality of the mask. In most cases, such interactions take the form of clicks, but scribbles, bounding boxes and coarse masks constitute usable forms of user guidance as well. 

The class agnostic nature of this task renders it viable for any kind of prompt. This property has been exploited to create a large foundation model which is capable of performing interactive segmentation, the Segment Anything model or SAM \citep{kirillov2023segment}. 
While SAM is trained on the large SA-1B dataset, which has been published in conjunction with the model, a lot of practical scenarios require the creation of datasets for very specific tasks. This is for example the case in smaller companies that seek to create datasets for the usage of in-house applications of computer vision, such as the automatization of processes. Here, only a small set of objects might be interesting to annotate. 
The go-to solution in such cases is fine-tuning the pretrained foundation model. Such a fine-tuning training, however, necessitates two factors: 
1) Availability of a preexisting dataset in the target domain that can be used as training data during fine-tuning.
2) The necessary computational resources to fine-tune the interactive segmentation model. This practically entails an entirely new additional training stage. On low-performance devices, such as hardware without GPU support or mobile phones, this requirement constitutes a considerable obstacle. 

Especially the latter problem occurs in situations where the annotation of data should be distributed amongst many annotators. Most of them will not have a high performance machine at their disposal. The goal is therefore to not only find a strategy for adapting an interactive segmentation model that does not require additional data, but one that is efficient in the sense that any computationally demanding fine-tuning process can be avoided completely. 

In our paper we are going to present such an adaptation strategy for SAM, while viewing this problem in the light of the interactive segmentation task on scenarios which are considerably different from regular consumer images. This first and foremost means the usage of appropriate metrics: The first important metric is the Number of Clicks (NoC) we need to annotated an object mask, and the second one is the Failure Rate (FR) which tells us about the percentage of cases in which we fail to do so with a reasonable number of clicks. Out of these two metrics, we regard the failure rate as the more crucial metric since it informs us about the limits of the model's segmentation capabilities. 
Our adaptation strategy mostly relies on pseudo-labels which are generated during the interaction. We use the clicks created by the user as pseudo labels for single pixels. In addition to that we use the mask which results from the interaction after pruning it to avoid errouneous training signals.

We will only carry out a partial adaptation of the network. In case the user intends to annotate multiple classes, the fine-tuned part can thus be copied for every particular class.
For the purpose of validating the techniques we are going to adapt SAM to miscellaneous rare situations, as well as medical image segmentation tasks. 
Our contributions can be summarized as follows: 
\begin{enumerate}
    \item We explore the performance of SAM as an interactive segmentation model on a variety of datasets which differ from regular consumer images.
    \item We test the limit of SAM's segmentation capabilities, and show that the model displays a considerable failure rate on domains which are different from general consumer images. 
    \item We show possible adaptation schemes which lower the failure rate without incurring considerable costs. The low memory overhead and fast adaptation render the usage of our method effectively for free.
\end{enumerate}

\section{Related Work}
\subsection{Interactive Segmentation} 
Interactive Segmentation uses various kinds of user guidance, with clicks being the most popular annotation mode \cite{chen2021conditional,  lin2020interactive, mahadevan2018iteratively, majumder2019content, xu2016deep}. 
The method in \cite{Man+18} uses four extreme points, which are assumed to be exactly on the borders of the object. Building on this work, \citet{dupont2021ucp} try to segment the object with only two non axis aligned points. 
\citet{jang2019interactive} try to improve their prediction by optimizing their interaction maps via backpropagation. The work of \citet{sofiiuk2020f} extends this by introducing auxiliary variables, which are optimized instead of the interaction maps. 
\citet{zhang2020interactive} combines bounding boxes with clicks on the object surface as user input. 
While recent work mostly uses on convolutional architectures \cite{ritm2022, chen2022focalclick, hao2021edgeflow}, the general training scheme is applied to networks with ViT-based backbones by \citet{liu2022simpleclick}. 
The methods in \cite{lowes2023interactive, chen2023scribbleseg, andriluka2020efficient, agustsson2019interactive, bai2014error} use scribbles as a form of guidance for interactive segmentation.
\cite{shi2022hybrid, liu2022isegformer} look at the problem of 3D interactive segmentation. Recently, \citet{kirillov2023segment} have proposed the so called Segment Anything model (SAM) together with SA-1B, the largest interactive segmentation dataset to date containing over 1.1B segmentation masks.
Due to the availability of the weights of the Segment Anything Model, there have been various papers which fine-tune its weights in order to adapt the model to a specific task. \citet{cheng2023sammed2d} and \citet{wu2023medical} adapt SAM to various medical image segmentation tasks. \citet{wang2023sam} use a modified version of SAM for robotic surgery. 
In \citet{chen2023sam}, adapter layers are introduced at intermediate places in the SAM-Encoder in order to fine-tune SAM to unusual image segmentation tasks. 
The method in \citet{ding2023adapting} adapts FastSAM \citep{zhao2023fast} for the task of change detection in remote sensing. The authors of \cite{sam_hq} improve SAM by adding a small amount of parameters to the SAM head and fine tuning these new parameters on high-quality human annotated data. It should be noted that all aforementioned methods require some additional fine-tuning on an existing annotated dataset in the target domain before they can be used. In contrast to that, our method can be used directly and adapts the network on-the-fly.

\subsection{Test-Time Adaptation}
The field of test-time adaptation deals with techniques to improve the model while it is already in use. Most of the existing methods are employed in contexts where there is no access to high-quality pseudo-labels, as would be the case in interactive segmentation. The methods proposed by \citet{song2023ecotta} and \citet{wang2020tent} leverage entropy-minimization to adapt the model. 
\citet{wang2022continual} use a consistency loss and a exponential moving average, while stochastically restoring single weights to mitigate error accumulation. 
The methods most strongly related to this paper, are methods which focus on the adaptation of interactive segmentation models during usage. The most commonly exploited information in these methods are the user generated clicks. Albeit very sparse, they provide immediately available ground truth information. \citet{kontogianni2020continuous}, \citet{shi2023self} and \citet{lenczner2020interactive} all exploit the clicks which are available due to the user interaction. The authors of \citet{wang2018interactive} fine-tune their model on the basis of scribbles. The works of \cite{hao2022rais} and \cite{lin2023sequential} is most similar to our method, since the authors mention that they use intermediate masks or previously created masks, respectively. They do, however, not mention any method avoiding erroneous masks or regions. In contrast to our method, both publications also introduce additional modules to their model which would require an additional previous fine-tuning stage.

\section{Method}
\subsection{Problem Statement}
\label{sec:problem}
First, we will provide a precise description of the interactive segmentation problem. We follow the problem description discussed in  \citep{ritm2022, liu2022simpleclick, chen2022focalclick}. Afterwards, we will briefly describe how we simulate the interaction in order to test such a system. Assume that we have an image $\bm{x} \in \mathbb{R}^{H \times W \times 3}$ and wish to create a segmentation map $\bm{m} \in \{0,1\}^{H,W}$ which delimits a desired area in said image. That is, every pixel belonging to the area in $\bm{x}$ is set to $1$ in $\bm{m}$, and every other pixel to $0$. 
In order to create such an annotation, a user will repeatedly interact with a neural network $f_\text{Seg}$ by providing it with clicks that indicate pixels reliably belonging to the foreground or background of the image. In each step $t$ the user will be shown the current estimation of the mask $\bm{m}_{t-1}$, which only consists of background pixels in the beginning ($t=0$). The user then chooses a falsely labeled region from the mask and places a click $\bm{p}_t$ on its surface. This $\bm{p}_t$ is a triple $(i_t, j_t, l_t)$ which indicates a position $(i, j) \in  \{1, ..., H\} \times \{1, ..., W\}$ and, depending on the choice of the user, a label $l \in \{+, -\}$ marking the position as foreground or background. The model $f_\text{Seg}$ is then given $\bm{m}_{t-1}$, all previously clicked pixels $\bm{p}_{1:t} = \{\bm{p}_1, ..., \bm{p}_t\}$ and the image $\bm{x}$ in order to predict an improved mask $\bm{m}_t = f_\text{Seg}(\bm{x}, \bm{p}_{1:t}, \bm{m}_{t-1})$. 
Once the user regards the quality of the mask as satisfactory, the interaction stops by saving this mask as $\bm{m}^\text{Res}$, and the next image is annotated. It is to be noted that this result mask $\bm{m}^\text{Res}$ might still be partially erroneous if the user chooses to ignore falsely annotated parts.

When it comes to evaluating the quality of such systems, we do not usually have a user at our disposal. Instead, we follow \citet{ritm2022} to simulate user interaction on images for which we already have ground truth segmentation masks $\bm{m}^\text{GT}$. At each iteration, we first compute the false positive area $\bm{m}_\text{FP}$ and the false negative area $\bm{m}_\text{FN}$. 
Then we compute the euclidean distance transforms $\mathcal{D}(\bm{m}_\text{FP})$ and $\mathcal{D}(\bm{m}_\text{FN})$ of the respective error masks, and select the pixel with the largest value on both distance transforms as a click. 
The label of the click depends on whether it has been placed on $\bm{m}_\text{FP}$ or $\bm{m}_\text{FN}$. We stop the interaction once the overlap of the proposed mask $\bm{m}_t$ with the ground truth mask $\bm{m}^\text{GT}$ exceeds a certain minimum IoU. This final mask will then be treated as the result mask $\bm{m}^\text{Res}$.

\subsection{Foundation models for Interactive Segmentation}
The Segment Anything Model (SAM) is a large foundation model for the general task of \emph{promptable segmentation}, which has been published in \citet{kirillov2023segment} alongside the SA-1B dataset. 
Promptable segmentation denotes the task of segmenting arbitrary object instances as indicated by a user interaction, such as bounding boxes, text prompts or foreground/background clicks, as well as previously available low-quality masks. The ability to improve upon previous masks and being guided by foreground/background clicks renders every promptable segmentation model compatible with click-based interactive segmentation. 
In addition to that, SAM has been pretrained on the SA-1B dataset, which contains 1.1B class-agnostic segmentation masks for 11M images. This causes SAM to be an extraordinarily good model for segmentation of objects on consumer images. Despite this, there is still room for improvement when it comes to more specific image domains and more obscure types of objects, as our experiments indicate. 

\begin{figure*}
    \centering
    \includegraphics[width=0.85\linewidth]{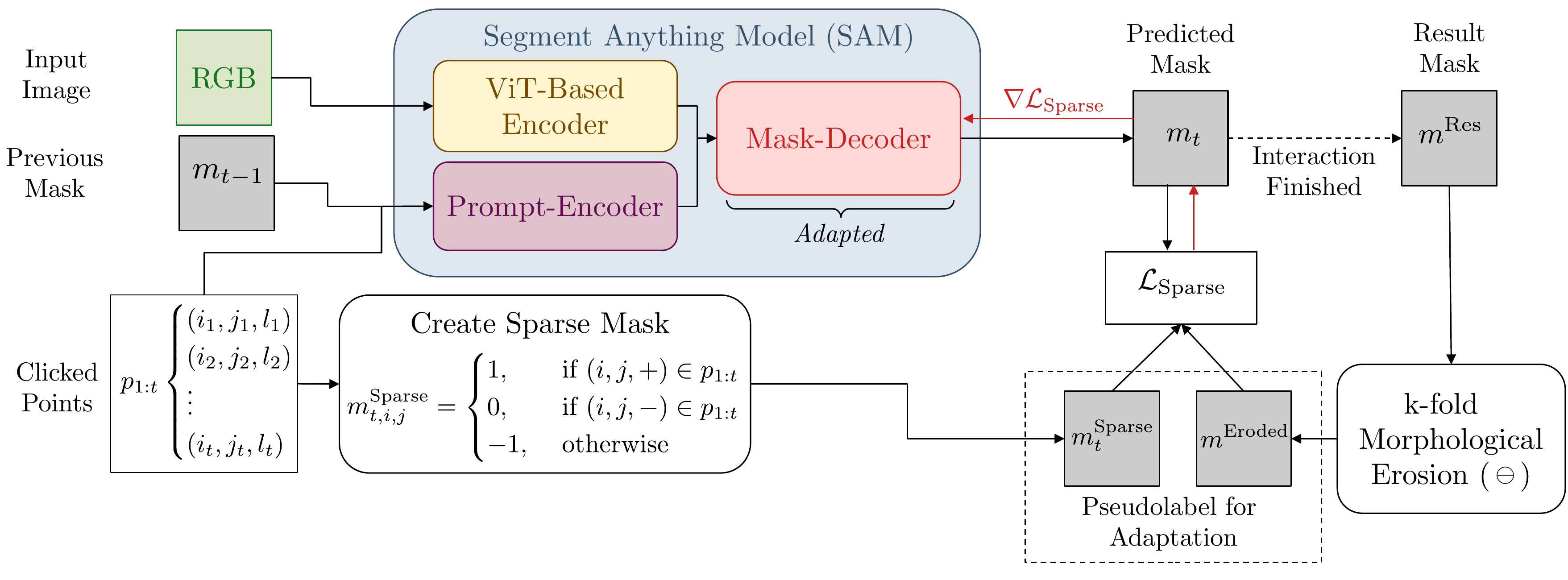}
    \caption{\label{fig:sam} A rough description of the SAM architecture and the information used as pseudo-labels. Our method only adapts the mask-decoder which renders the computational effort of the backpropagation and optimization negligible. The gradient computation is displayed in {\color[HTML]{c9211e} red}. The usage of pseudo-labels is discussed in  \Cref{sec:adaption}.}

\end{figure*}

The architecture of SAM itself is divided into three parts: An \emph{image encoder}, a \emph{prompt encoder} and a \emph{mask decoder}. 
The image encoder receives an image $\bm{x} \in \mathbb{R}^{H \times W \times 3}$ and encodes it into a feature map independently of any user interaction. The authors of SAM use a ViT backbone for this task. 
The prompt encoder receives the prompt in the form of clicks, bounding boxes, and masks, and encodes them into a form which is useful for the mask decoder. 
The mask decoder receives the image features and the encoded prompts, and uses both to predict as segmentation mask for the object indicated by the prompts. \Cref{fig:sam} contains a rough visualization of the SAM architecture.

The greatest benefit of this general architecture lies in the decoupling of the computation of prompt embeddings and image features. The image only needs to be embedded once, while additional interactions only require a reuse of the prompt encoder and mask decoder. As long as the latter two networks are sufficiently light-weight, the user will be granted a real-time experience during the interactive usage of the model. 

\subsection{Adapting the Model During Test-Time }
\label{sec:adaption}
When performing interactive segmentation, we generally annotate a sequence of images instead of just a single one. This opens up the possibility of exploiting information gathered from segmenting previous images, in order to get better at segmenting future images. 
Similar to \citet{kontogianni2020continuous} and \citet{lenczner2020interactive}, we make use of the fact that each click on its own constitutes a single reliably correct ground truth pixel. Since this piece of ground truth is available directly after being entered by the user, we can already adapt the model while still annotating the image. 
Additionally, we use the mask $\bm{m}^\text{Res}$ which results after the user is done annotating the image. Depending on the users judgement, some areas of $\bm{m}^\text{Res}$ may still be erroneous. Since we especially suspect the borders between foreground and background to be faulty, we first subject the mask to multiple iterations of morphological erosion and then use this eroded mask $\bm{m}^\text{Eroded}$ as a pseudo-label to adapt the model to the image domain.
When carrying out the adaptation, we only optimize the parameters of the decoder. 
A single execution of backpropagation and optimization with the Adam optimizer took 43.6 ms on a Nvidia V100 GPU vs. 13.1 ms for the corresponding forward pass. 
Since the accompanying optimization takes less than four times the time of the forward pass, the method doesn't impede any potential real time usage. Extracting the features with the backbone takes 116.9 ms. This operation, however, only has to be executed once per image.
In the following paragraphs, we describe the variants of adaptation used by us.

\paragraph{Immediately using Clicks for Adaptation. } As soon as the user makes a click $\bm{p}_t = (i_t, j_t, l_t)$, we have ground truth information for a particular pixel at our disposal. We can use all clicks $\bm{p}_{1:t}$ we have received up until that point in order to create a sparse mask $\bm{m}^\text{Sparse}_t$ with 
\begin{equation}
    \label{eq:sparsemask}
    \bm{m}^\text{Sparse}_{t, i, j} = 
    \begin{cases}
        1, &\text{ if } (i, j, +) \in \bm{p}_{1:t}\\ 
        0, &\text{ if } (i, j, -) \in \bm{p}_{1:t}\\
        -1, &\text{ otherwise } 
    \end{cases}
\end{equation}
where $-1$ marks unknown pixels. Let $\bm{m}_t$ be the segmentation mask that has been computed after that last click has been made.  We then compute a sparse binary cross entropy loss 
\begin{equation}
\begin{split}
    \mathcal{L}_\text{Sparse}(\bm{m}^\text{Sparse}_t, \bm{m}_t) &= \frac{\sum_{i, j} 1_{\bm{m}^\text{Sparse}_{t, i, j} = 1} \mathcal{L}_\text{BCE}(\bm{m}^\text{Sparse}_{t, i, j}, m_{t,i, j})}{\sum_{i, j} 1_{\bm{m}^\text{Sparse}_{t, i, j} = 1}} \\
    &+ \frac{\sum_{i, j} 1_{\bm{m}^\text{Sparse}_{t, i, j} = 0} \mathcal{L}_\text{BCE}(\bm{m}^\text{Sparse}_{t, i, j}, m_{t, i, j})}{\sum_{x,y} 1_{\bm{m}^\text{Sparse}_{t, i, j} = 0}}
\end{split}
\end{equation}
using $\bm{m}^\text{Sparse}_t$ as the label mask. We then immediately carry out an optimization step, thus progressively overfitting to the particular image as we continue annotating it. Note that this overfitting is deliberate and has to be reversed after we are done with the image. In order to achieve this, we reset the weights to their values before the image annotation, directly after we are done with the image.

\paragraph{Using all Clicks to adapt the Model to the Image Sequence. } While the last paragraph describes a deliberate overfitting to the image, we also have the option to only carry out a single optimization step after we finish annotating the image. When doing this, we use all clicks that have been accumulated during the annotation of an image to create a single $\bm{m}^\text{Sparse}$ per image. The mask is created in the same fashion as before. This strategy adapts the model to the type of object and image domain of the test set, whilst acting less destructive on the parameters.

\paragraph{Using the Resulting Mask to Adapt the Model to the Image Sequence. } Once the user regards the interactively created mask to be of sufficient quality, they stop the annotation and we obtain the result mask $\bm{m}^\text{Res} \in \{0, 1\}^{H \times W}$. We can use this mask as a pseudo-label to adapt the model to the image sequence.  
In order to circumvent erroneous regions we will prune $\bm{m}^\text{Res}$ at the borders between foreground and background, where we estimate the risk of errors to be the highest. This is done by separating the foreground and background masks, iteratively eroding both of them and uniting them again. Let $\bm{m}^\text{FG} = \bm{m}^\text{Res}$ and $\bm{m}^\text{BG} = 1 - \bm{m}^\text{Res}$ be the foreground and background masks, respectively. We define $\gamma^k(\bm{m})$ to be a $k$-fold application of morphological erosion as  
\begin{align}
    \gamma^0(\bm{m}) &= \bm{m}, \\
    \gamma^{k}(\bm{m}) &= \gamma^{k-1}(\bm{m}) \ominus \begin{bmatrix}
        0 & 1 & 0 \\
        1 & 1 & 1 \\ 
        0 & 1 & 0
    \end{bmatrix}, 
\end{align}
where $\ominus$ is the symbol for the erosion operation. Then $\bm{m}^{\text{FG, Eroded}} = \gamma^k(\bm{m}^\text{FG})$ and $\bm{m}^{\text{BG, Eroded}} = \gamma^k(\bm{m}^\text{BG})$ are the eroded background and foreground masks. We will unite the two, resulting in the pruned pseudolabel mask $\bm{m}^\text{Eroded}$ with 
\begin{equation}
    m^\text{Eroded}_{i,j} = 
    \begin{cases}
        1, & \text{ if } m^{\text{FG, Eroded}}_{i,j} = 1 \\
        0, & \text{ if } m^{\text{BG, Eroded}}_{i,j} = 1 \\
        -1, & \text{ otherwise}
    \end{cases}. 
\end{equation}
We will carry out a single optimization step using $\mathcal{L}_\text{Sparse}$ after annotating each image.

\paragraph{Using multiple decoders for Multiple Classes.}
All of the aforementioned adaptation will inevitably overfit the model to a particular domain or type of object. 
It is however noteworthy, that the only part of the model to be adapted is the relatively lightweight decoder. 
This allows use to duplicate the parameters of the adapted module. In cases where we want to annotate multiple different classes, we use multiple copies of the original decoder, which are separately adapted to the respective object type or domain. 
We regard the memory overhead as negligible: 
For the version of SAM with the ViT-b backbone, we have 4.06M parameters for the decoder vs 89.7M parameters for the rest of the model. For the versions with the ViT-l and ViT-h backbones, the rest of the model has 308.3M and 637M parameters respectively, while the decoder size remains the same.

\section{Experiments} 

\begin{figure*}
    \centering
    {\def\plotheight{0.09\linewidth}
    \setlength{\tabcolsep}{0.5pt}
    \begin{tabular}{ccccccc} 
         & CAMO & TimberSeg & LeafDisease & KvasirSeg & KvasirInstrument  & PPDLS \\
         \rotatebox{90}{\quad \; \textbf{GT}} & \includegraphics[height=\plotheight]{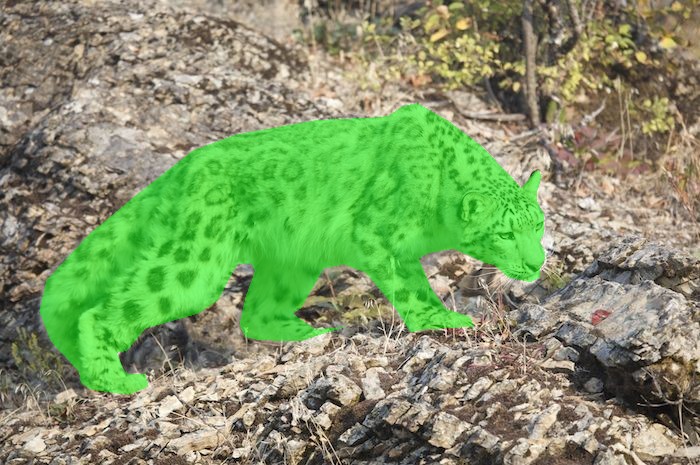} & \includegraphics[height=\plotheight]{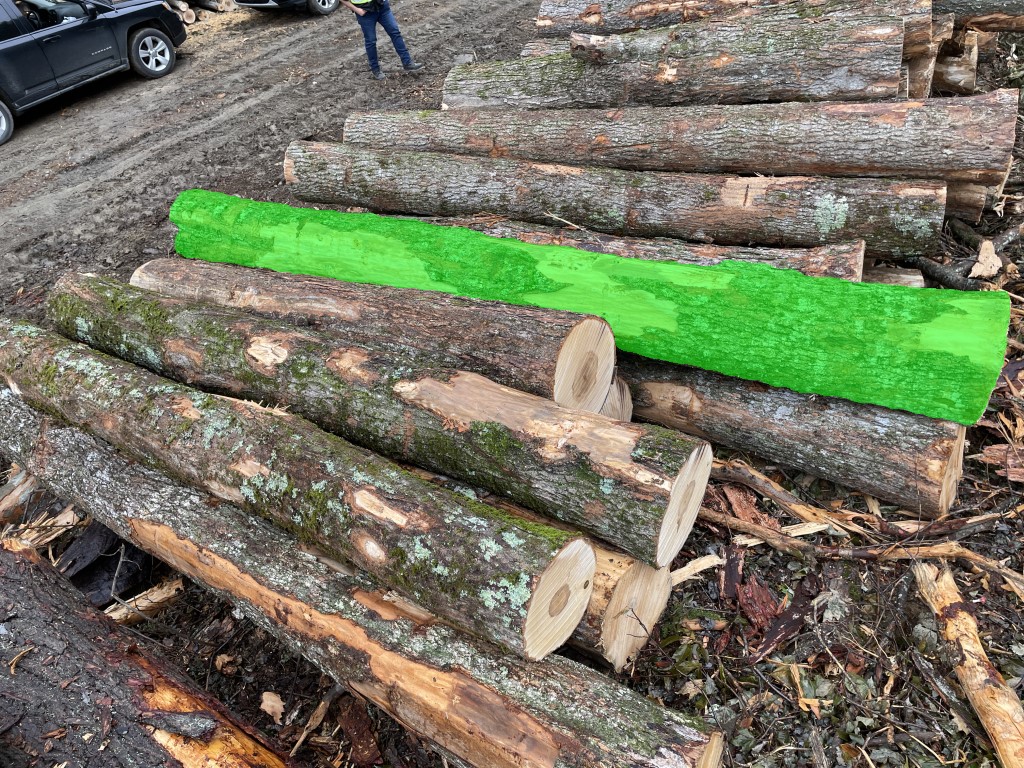} & \includegraphics[height=\plotheight]{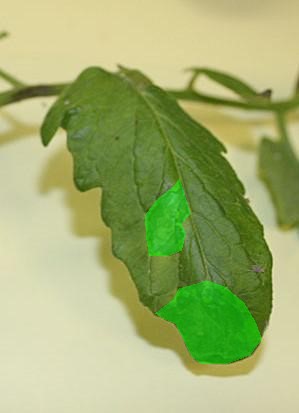} & \includegraphics[height=\plotheight]{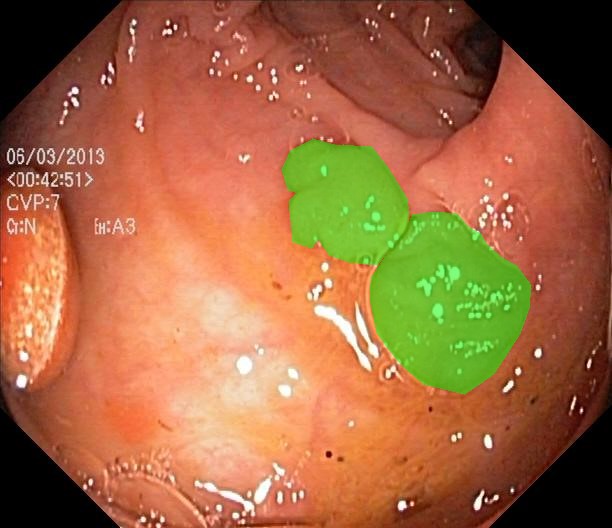} & \includegraphics[height=\plotheight]{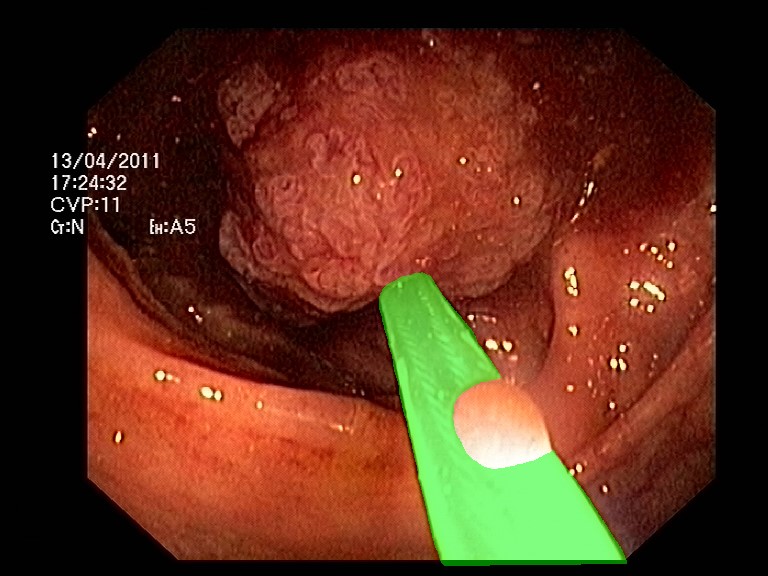} & \includegraphics[height=\plotheight]{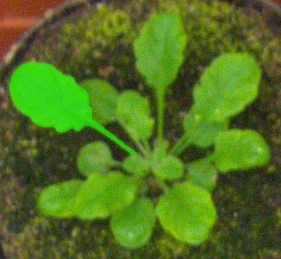}\\
         
         \rotatebox{90}{\,\textbf{Clicks}} & \includegraphics[height=\plotheight]{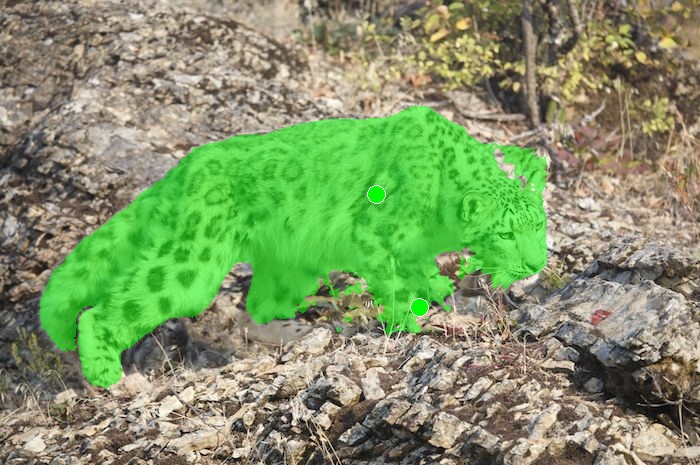} & \includegraphics[height=\plotheight]{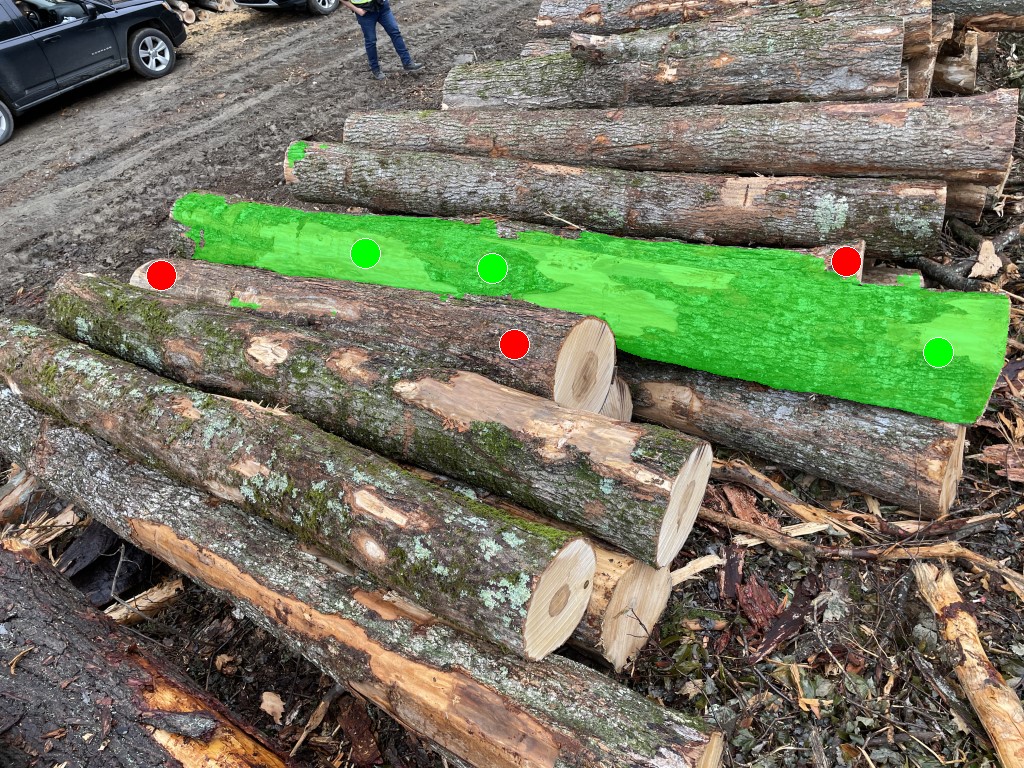} & \includegraphics[height=\plotheight]{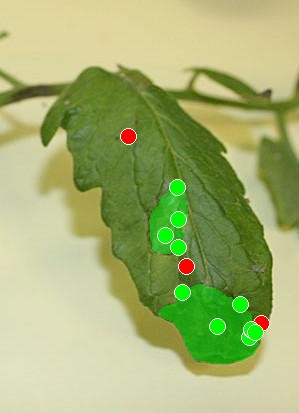} & \includegraphics[height=\plotheight]{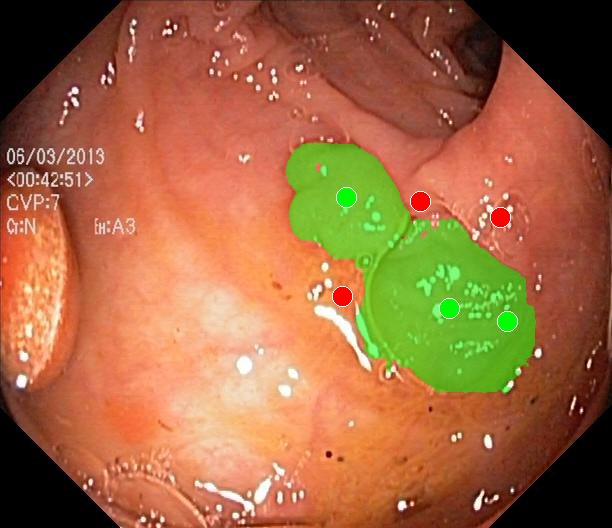} & \includegraphics[height=\plotheight]{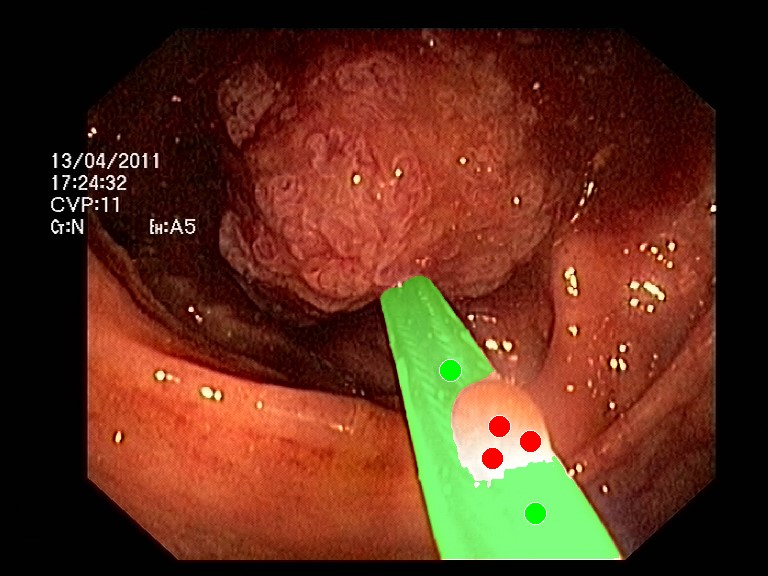} & \includegraphics[height=\plotheight]{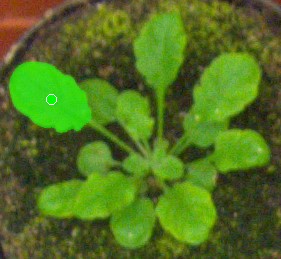}\\
         
         \rotatebox{90}{\; \textbf{Eroded}} & \includegraphics[height=\plotheight]{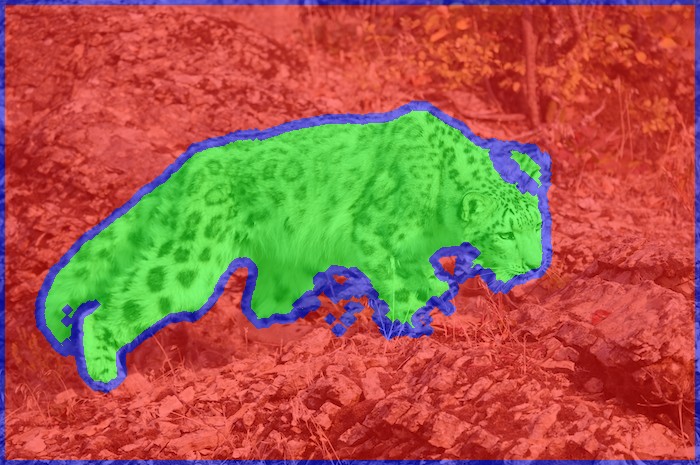} & \includegraphics[height=\plotheight]{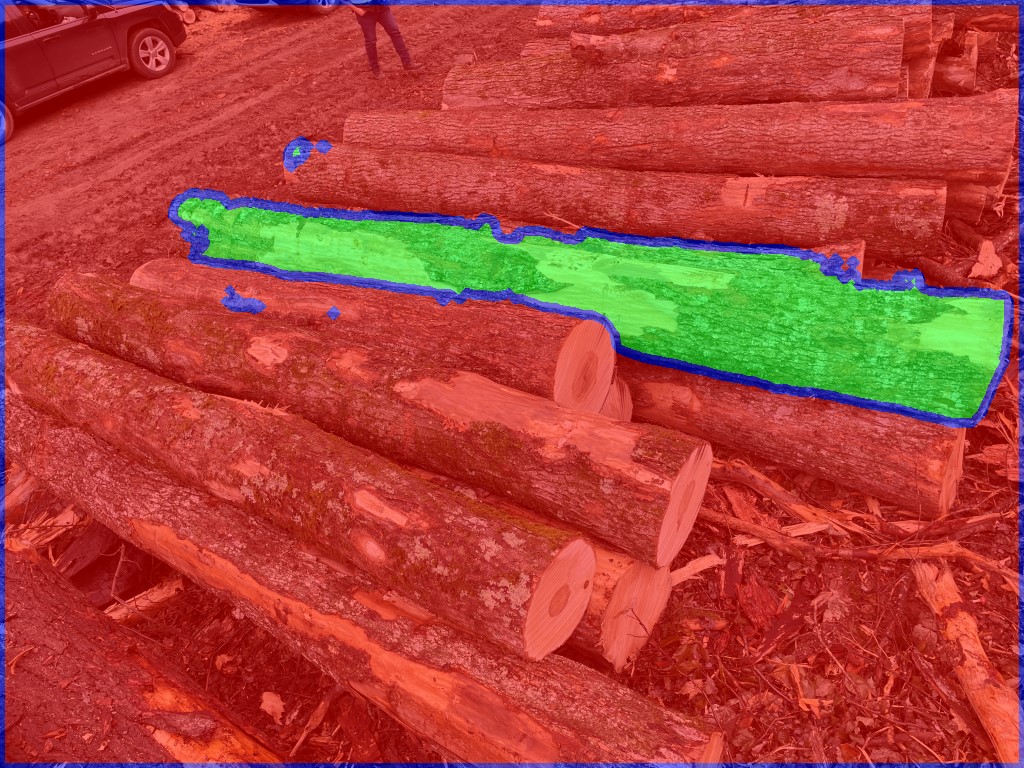} & \includegraphics[height=\plotheight]{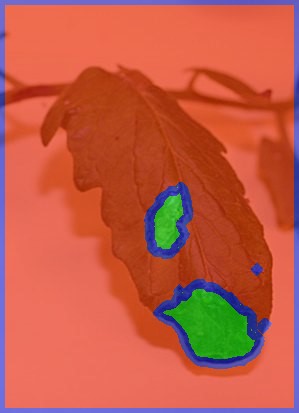} & \includegraphics[height=\plotheight]{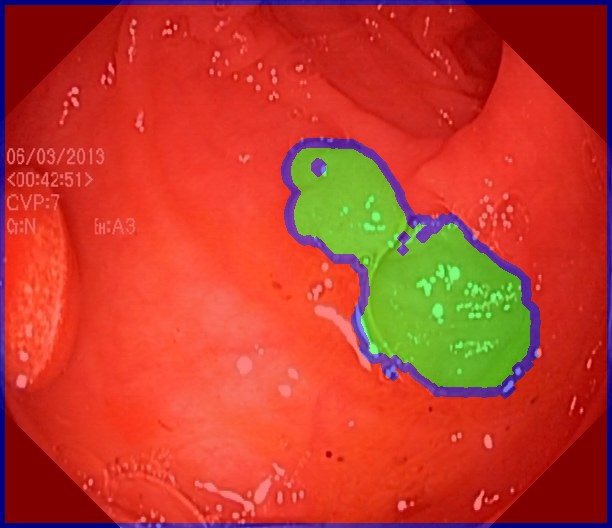} & \includegraphics[height=\plotheight]{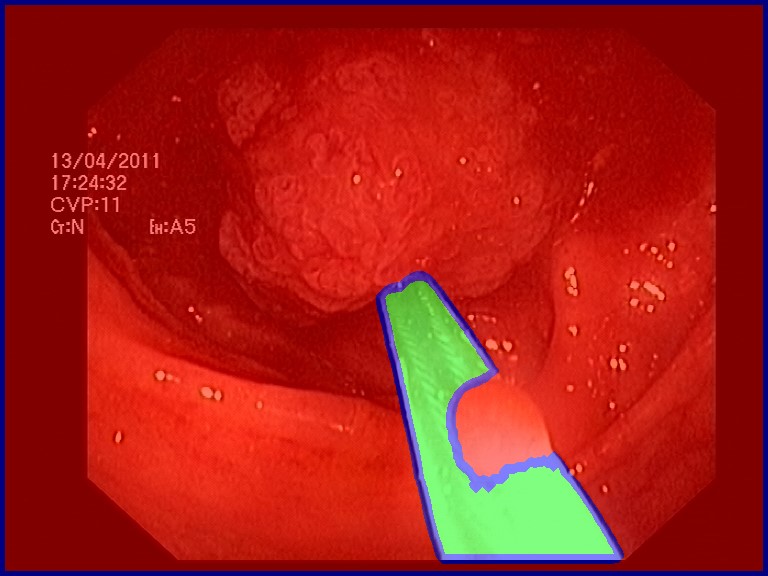} & \includegraphics[height=\plotheight]{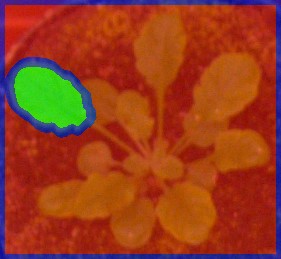} 
    \end{tabular}
    }
    \caption{Examples for the masks occurring during the interaction. The \emph{first} row contains the ground truth. The \emph{second} row contains the annotated mask and the clicks. The \emph{third} row contains examples for the eroded result mask. Green, red and blue correspond to foreground, background and the eroded area, respectively. \label{fig:qualitative}}
    
\end{figure*}

\begin{table*}[t]
    \centering
    \begin{tabular}{|c|c|c||c|c|c|c|c|c|c|c|c|c|}
        \hline
        \multicolumn{3}{|c||}{Configuration} & \multicolumn{2}{|c|}{Rooftop} & \multicolumn{2}{|c|}{DOORS}  & \multicolumn{2}{|c|}{TrashCan}   & \multicolumn{2}{|c|}{CAMO} \\
        \hline
         CA & RM & CM & NoC & FR & NoC & FR & NoC & FR & NoC & FR \\
         \hline
         & & & 4.171 & 6.00 & 5.439 & 16.69 & 13.259 & 57.42 & 7.224 & 20.3 \\ 
         \hline
         R & E & $\checkmark$ & 3.667 & 3.93 & 4.877 & 13.50 & 11.488 & 40.49 & 7.310 & 17.2\\
         \hline
         R & & & 3.755 & 3.93 & 5.149 & 12.25 & 11.847 & 39.41 & 7.382 & 18.2 \\
         C & & & 3.834 & 3.93 & 5.222 & 12.73 & 11.932 & 41.42 & 7.212 & 17.1 \\
         & E & & 3.741 & 3.39 & 5.642 & 18.10 & 13.486 & 58.23 & 7.401 & 20.2 \\
         & & $\checkmark$ & 3.915 & 4.62 & 5.154 & 14.97 & 13.694 & 59.47 & 7.278 & 19.4 \\ 
         R & & $\checkmark$ & 3.707 & 3.70 & 5.326 & 12.83 & 11.796 & 40.38 & 7.402 & 17.0 \\ 
         R & U & $\checkmark$ & 3.693 & 3.00 & 4.861 & 12.64 & 16.041 & 64.49 & 12.764 & 45.8 \\
         \hline
         \multicolumn{3}{|c||}{HQ-SAM} & 9.977 & 31.64 & 10.688 & 42.74 & 16.902 & 79.83 & 10.383 & 36.5 \\
         \hline

        \hline         
        \multicolumn{3}{|c||}{Configuration} & \multicolumn{2}{|c|}{ISTD} & \multicolumn{2}{|c|}{LeafDisease} & \multicolumn{2}{|c|}{PPDLS} & \multicolumn{2}{|c|}{TimberSeg} \\
        \hline
         CA & RM & CM & NoC & FR & NoC & FR & NoC & FR & NoC & FR \\
         \hline
         & & &  11.584 & 40.68 & 14.624 & 62.07 & 6.239 & 23.76 & 11.564 & 48.50 \\ 
         \hline
         R & E & $\checkmark$ & 10.392 & 31.13 & 14.595 & 60.71 & 6.250 & 20.04 & 10.497 & 39.67 \\
         \hline
         R & & & 10.932 & 34.66 & 14.665 & 61.05 & 6.267 & 19.25 & 11.080 & 42.26 \\
         C & & & 10.896 & 33.91 & 14.631 & 60.71 & 6.218 & 19.43 & 10.661 & 40.73 \\
         & E & & 11.295 & 38.80 & 14.690 & 61.05 & 5.955 & 21.42 & 10.745 & 43.32 \\
         & & $\checkmark$ & 11.596 & 41.73 & 14.517 & 60.54 & 5.988 & 21.56 & 10.933 & 43.92 \\ 
         R & & $\checkmark$ & 10.810 & 33.68 & 14.469 & 60.03 & 6.140 & 19.54 & 10.571 & 40.18 \\ 
         R & U & $\checkmark$ & 15.017 & 57.97 & 14.918 & 62.41 & 14.387 & 49.40 & 16.710 & 74.76 \\
         \hline
         \multicolumn{3}{|c||}{HQ-SAM} & 18.757 & 89.32 & 16.519 & 74.49 & 10.173 & 3646 & 17.706 & 84.33 \\
         \hline
    \end{tabular}
    \caption{The results on datasets displaying rare objects types. NoC means the $\text{NoC}_{20}@85$ metric and FR is the $\text{FR}_{20}@85$, describing the number of objects that could not be segmented after 20 clicks. For both metrics, a smaller value indicates a better performance. An explanation of the configurations can be found in \Cref{sec:setting}. \label{tab:obscure85} } 
    
\end{table*}

\begin{table*}
    \centering 

    \begin{tabular}{|c|c|c||c|c|c|c|c|c|c|c|c|c|}
        \hline
        \multicolumn{3}{|c||}{Configuration} & \multicolumn{2}{|c|}{Rooftop} & \multicolumn{2}{|c|}{DOORS} & \multicolumn{2}{|c|}{TrashCan} & \multicolumn{2}{|c|}{CAMO} \\
        \hline
         CA & RM & CM & NoC & FR & NoC & FR & NoC & FR & NoC & FR \\
         \hline 
         & & & 9.979 & 22.63 & 13.870 & 37.77 & 23.281 & 72.49 & 13.870 & 34.1 \\ 
         \hline
         R & E & $\checkmark$ & 8.891 & 18.21 & 13.163 & 33.62 & 20.527 & 54.06 & 13.488 & 28.3 \\
         \hline
         R & & & 8.961 & 18.24 & 14.996 & 36.30 & 20.979 & 53.86 & 13.719 & 29.6 \\
         C & & & 9.358 & 19.86 & 14.623 & 35.35 & 21.032 & 53.40 & 13.573 & 29.1 \\
         & E & & 9.321 & 19.63 & 14.965 & 42.47 & 23.700 & 73.30 & 14.082 & 33.0 \\
         & & $\checkmark$ & 9.314 & 19.40 & 13.629 & 35.96 & 23.976 & 74.27 & 14.063 & 33.6 \\ 
         R & & $\checkmark$ & 9.127 & 18.94 & 15.533 & 37.33 & 20.925 & 52.20 & 13.503 & 28.5 \\ 
         R & U & $\checkmark$ & 9.339 & 19.40 & 13.082 & 33.31 & 25.221 & 70.75 & 20.840 & 54.2 \\
         \hline
         \multicolumn{3}{|c||}{HQ-SAM} & 19.637 & 53.12 & 20.475 & 61.10 & 26.844 & 87.09 & 18.010 & 50.0  \\
         \hline

        \hline        
        \multicolumn{3}{|c||}{Configuration} & \multicolumn{2}{|c|}{ISTD} & \multicolumn{2}{|c|}{LeafDisease} & \multicolumn{2}{|c|}{PPDLS} & \multicolumn{2}{|c|}{TimberSeg} \\
        \hline
         CA & RM & CM & NoC & FR & NoC & FR & NoC & FR & NoC & FR \\
         \hline
         & & & 18.744 & 49.02 & 24.255 & 72.62 & 13.260 & 38.55 & 20.358 & 62.64 \\ 
         \hline 
         R & E & $\checkmark$ & 16.660 & 40.00 & 23.617 & 70.24 & 13.782 & 30.28 & 18.735 & 52.15 \\
         \hline
         R & & & 17.411 & 41.80 & 24.138 & 71.26 & 13.682 & 31.30 & 19.018 & 54.46 \\
         C & & & 17.302 & 40.90 & 24.214 & 72.28 & 13.276 & 30.88 & 19.026 & 54.00 \\
         & E & & 18.329 & 47.89 & 24.320 & 72.62 & 12.877 & 36.17 & 19.306 & 58.21 \\
         & & $\checkmark$ & 19.574 & 53.08 & 24.226 & 71.60 & 12.574 & 35.07 & 19.436 & 58.76 \\ 
         R & & $\checkmark$ & 17.217 & 41.35 & 24.153 & 72.11 & 13.447 & 31.22 & 18.874 & 53.49 \\ 
         R & U & $\checkmark$ & 22.729 & 59.40 & 24.221 & 72.11 & 22.892 & 56.13 & 26.319 & 79.89 \\
         \hline
         \multicolumn{3}{|c||}{HQ-SAM} & 28.337 & 91.20 & 26.269 & 81.63 & 18.180 & 49.87 & 27.364 & 88.58  \\
         \hline
    \end{tabular}
    
    \caption{The results on datasets displaying rare objects types. NoC means the $\text{NoC}_{30}@90$ metric and FR is the $\text{FR}_{30}@90$, describing the number of objects that could not be segmented after 30 clicks. For both metrics, a smaller value indicates a better performance.  An explanation of the configurations can be found in \Cref{sec:setting}. \label{tab:obscure90} }
\end{table*}

\begin{table*}
    \centering
    
    \begin{tabular}{|c|c|c||c|c|c|c|c|c|c|c|}
        \hline
        \multicolumn{3}{|c||}{Configuration} & \multicolumn{2}{|c|}{KvasirInstrument} & \multicolumn{2}{|c|}{CVCClinicDB} & \multicolumn{2}{|c|}{GlaS} & \multicolumn{2}{|c|}{KvasirSeg} \\
        \hline
         CA & RM & CM & NoC & FR & NoC & FR & NoC & FR & NoC & FR\\
         \hline
         & & & 2.137 & 1.86 & 4.935 & 8.17 & 7.485 & 14.64 & 3.615 & 2.7 \\ 
         \hline 
         R & E & $\checkmark$ & 2.166 & 1.53 & 4.551 & 5.56 & 6.759 & 10.20 & 3.145 & 1.4 \\
         \hline
         R & & & 2.388 & 2.71 & 4.828 & 5.39 & 7.377 & 13.53 & 3.314 & 1.1 \\
         C & & & 2.239 & 2.37 & 4.900 & 7.03 & 7.250 & 13.27 & 3.352 & 1.2 \\
         & E & & 2.136 & 1.69 & 4.471 & 4.41 & 8.437 & 20.65 & 3.123 & 1.2\\
         & & $\checkmark$ & 2.178 & 2.37 & 4.637 & 5.39 & 8.539 & 20.72 & 3.281 & 1.2 \\ 
         R & & $\checkmark$ & 2.305 & 2.37 & 4.757 & 6.21 & 7.576 & 15.29 & 3.273 & 1.0 \\ 
         R & U & $\checkmark$ & 2.251 & 2.20 & 5.087 & 6.70 & 13.946 & 49.15 & 7.684 & 20.3 \\
         \hline
         \multicolumn{3}{|c||}{HQ-SAM} & 7.973 & 18.31 & 15.789 & 66.01 & 18.845 & 88.89 & 10.504 & 34.1 \\
         \hline
    \end{tabular}
    
    \caption{\label{tab:medical85} The results medical datasets. NoC means the $\text{NoC}_{20}@85$ metric and FR is the $\text{FR}_{20}@85$, describing the number of objects that could not be segmented after 20 clicks. For both metrics, a smaller value indicates a better performance. An explanation of the configurations can be found in \Cref{sec:setting}.}
\end{table*}

\begin{table*}
    \centering
    
    \begin{tabular}{|c|c|c||c|c|c|c|c|c|c|c|}
        \hline
        \multicolumn{3}{|c||}{Configuration} & \multicolumn{2}{|c|}{KvasirInstrument} & \multicolumn{2}{|c|}{CVCClinicDB} & \multicolumn{2}{|c|}{GlaS} & \multicolumn{2}{|c|}{KvasirSeg} \\
        \hline
         CA & RM & CM & NoC & FR & NoC & FR & NoC & FR & NoC & FR\\
         \hline
         & & & 3.651 & 4.75 & 10.301 & 19.61 & 14.995 & 33.53 & 6.378 & 5.8 \\ 
         \hline 
         R & E & $\checkmark$ & 3.825 & 4.58 & 8.585 & 10.46 & 11.684 & 19.15 & 5.580 & 3.9 \\
         \hline
         R & & & 4.063 & 5.42 & 9.343 & 14.05 & 13.341 & 24.12 & 6.397 & 5.7 \\
         C & & & 4.041 & 5.42 & 9.041 & 12.75 & 13.331 & 23.73 & 6.057 & 4.4 \\
         & E & & 3.749 & 5.08 & 9.588 & 14.87 & 15.884 & 35.49 & 5.573 & 3.4 \\
         & & $\checkmark$ & 3.647 & 4.75 & 9.458 & 14.87 & 16.729 & 40.13 & 6.106 & 4.9 \\ 
         R & & $\checkmark$ & 4.237 & 5.93 & 9.253 & 13.40 & 13.690 & 25.23 & 6.178 & 5.7 \\ 
         R & U & $\checkmark$ & 4.239 & 5.76 & 12.446 & 21.57 & 22.744 & 55.29 & 16.168 & 34.2\\
         \hline
         \multicolumn{3}{|c||}{HQ-SAM} & 13.698 & 30.85 & 24.139 & 70.75 & 28.888 & 93.86 & 17.410 & 44.4 \\
         \hline
    \end{tabular}
    \caption{The results medical datasets. NoC means the $\text{NoC}_{30}@90$ metric and FR is the $\text{FR}_{30}@90$, describing the number of objects that could not be segmented after 30 clicks. For both metrics, a smaller value indicates a better performance. An explanation of the configurations can be found in \Cref{sec:setting}. \label{tab:medical90} }
\end{table*}

\subsection{Experimental Setting }
\label{sec:setting}
\paragraph{Implementation Details.} During training we only adapt the decoder in order to minimize the computational overhead of our method. We carry out all optimization with a sparse binary cross entropy loss, as described in \Cref{sec:adaption}. 
We use the Adam optimizer \citep{kingma2014adam} with a learning rate of  $10^{-6}$. The resolution of the input images is $1024 \times 1024$, which is a pre-existing property of SAM. All experiments use the ViT-b backbone \citep{dosovitskiy2020image}. Whenever we use erosion, we carry out the iterative erosion with $k=5$ iterations.

\paragraph{Metrics.} 
When testing an interactive segmentation system, we want to exceed a certain IoU threshold $\mathtt{T}_\text{IoU}$ within $n$ clicks. If the system is unable to do that, we consider the attempt at segmenting the image a failure and use $n$ as surrogate value for the number of clicks when computing the $\text{NoC}_n@\mathtt{T}_\text{IoU}$. The \emph{Number of Clicks} ($\text{NoC}_n@\mathtt{T}_\text{IoU}$) metric measures the average number of clicks on the test set, while the \emph{Failure Rate} ($\text{FR}_n@\mathtt{T}_\text{IoU}$) measures the percentage of images on which the segmentation failed.
Out of the two metrics we regard the failure rate as the more important one for the following reason: While having to add an additional click on some images during the annotation process incurs a higher time effort, the failure rate measures the amount of images that cannot be segmented within a reasonable number of clicks at all. 

\textbf{Click Adaptation (CA): } After each click, we can use all so far accumulated clicks to create a sparse label mask, with which we optimize the model to overfit to the image. We call this process \emph{Click Adaptation (CA)}. 
In \Cref{sec:adaption} we mentioned that this deliberate overfitting necessitates resetting the weight after each object, which we denote with an R for (R)eset in the tables. We may however choose to not perform this reset, and adapt our model continually over all images. We denote this by a C for (C)ontinual. No letter in the tables means that we do not use Click Adaptation at all. 

\textbf{Result Mask (RM): } After being done with annotating an image, we can make use of the \emph{Result Mask (RM)}. We could directly use the mask as a pseudolabel for optimization. We denote this with a U for (U)ntreated in the tables. As we will show however, this mask may still be erroneous and worsen our performance by subjecting our model to a partially false training signal. In order to circumvent this problem we may prune the masks foreground and background area by using iterative erosion. We denote this by an E for (E)rosion. No letter means that we do not make use of the result mask. 

\textbf{Click Mask (CM): } After the annotation, we can use the accumulated clicks to form a sparse \emph{Click Mask (CM)}, with which we can perform a single optimization step. In each configuration in which we do so, it is annotated by a checkmark ($\checkmark$).  

The table row containing no letter or checkmark means that we are not performing any form of adaptation, which constitutes our baseline. Whenever we use the Result Mask and the Click Mask in the same configuration, we merge them into a single mask. 
In all tables, the first line contains the baseline, while the second line contains our complete method. \Cref{fig:qualitative} shows some qualitative examples.

\subsection{Adaptation to Rare Objects}

We will adapt SAM during usage on various datasets providing examples for rather obscure and uncommon situations. The Rooftop dataset \citep{datasetrooftop} provides various remote sensing photos with annotated rooftops. The DOORS dataset \citep{datasetdoors} has been created for the segmentation of boulders. The TrashCan dataset \citep{datasettrashcan} contains segmentation masks for underwater waste objects. CAMO \citep{datasetcamo1, datasetcamo2} is a dataset for the task of camouflaged object segmentation and ISTD \citep{datasetistd} for shadow segmentation. Additionally, we have three datasets for agricultural applications: One dataset for leaf disease segmentation \citep{datasetleafdisease}, PPDLS \citep{datasetppdls} for the segmentation of arabidopsis and tobacco leafs, and TimberSeg \citep{datasettimberseg} for the segmentation of logs in forestry work. 

We are first going to look at $\text{NoC}_{20}@85$ and $\text{FR}_{20}@85$ metrics. According to \Cref{tab:obscure85}, our method reduces the FR on ISTD from 40.68 to 31.13, while reducing the NoC by more than one click. On TrashCan, our method even improves the FR from 57.42 to 40.49. It should also be noted that the results imply that SAM is unable to segment over half of the objects in the TrashCan and LeafDisease datasets to a satisfying degree. While our complete method slightly increases the NoC on the CAMO and PPDLS datasets, it still lowers the FR which we regard as the more crucial metric. In order to see the effect of using the untreated mask, we also run a version of our complete method without pruning the mask by erosion. As it turns out, eroding the mask is important due to potential erroneous areas at the edge of foreground and background area. The resulting false training signal manages to increase the FR by even more than two times on CAMO. 

In \Cref{tab:obscure90}, where the model needs to achieve an IoU of 90 within 30 clicks, we see an exacerbation of the problem SAM has with segmenting objects that are alien to its original training set. The FR values of the unadapted SAM model are 72.49, 72.62 and 62.64 on TrashCan, LeafDisease, and TimberSeg, respectively. This indicates that SAM is almost inept to segment these types of data to an IoU of 90 with the actual object surface, which would be considered necessary when producing annotations for new data. In the case of TrashCan and TimberSeg we manage to reduce the FR by 18.43 and 10.49 percentage points, respectively. The largest improvements regarding the NoC are incurred on TrashCan with a reduction of 2.754 clicks. On PPDLS, we again have a reduction in the FR for the cost of slightly higher NoC. It should be noted, that our complete method (CA = R, RM = E, CM = $\checkmark$) reduces the failure rate in all cases, and thus widens the applicability of SAM for uncommon domains.

\subsection{Results on Medical Image Segmentation}
In order to investigate the efficacy of the adaptation method on medical image segmentation, we consider four different datasets: KvasirInstrument \citep{datasetkvasirinst} contains segmented images of tools used in the gastrointestinal tract. CVCClinicDB \citep{datasetcvcclinicdb} and KvasirSeg \citep{datasetkvasirseg} are two datasets for the task of polyp segmentation, while the GlaS dataset \citep{datasetglas1, datasetglas2} provides data for the task of gland segmentation in colon histology. 
The results for using our method on medical data generally comport with the results on other rare objects. It should first be noted that our complete method causes a reduction of the failure rate in all cases. In \Cref{tab:medical85} we see the complete method decreasing the FR on KvasirSeg from 2.7 to 1.4, which is a relative reduction of $48.1 \%$. On GlaS, the FR is lowered from 14.64 to 10.20 and the NoC is lowered from 7.485 to 6.759.
On KvasirSeg and GlaS, the untreated result mask with a partially erroneous signal causes the most damage. It increases the failure rate by 18.9 and 38.95 percentage points in comparison to the full method with the eroded mask on each of the respective datasets. 
In \Cref{tab:medical90}, we can see a reduction in the FR by 14.38 percentage points, as well as a reduction in the NoC by 3.311 clicks on GlaS. On CVCClinicDB the FR is lowered by 9.15 percentage points, which equates to a reduction of $46.6 \%$, while the NoC is lowered by 1.716 clicks.
On KvasirInstrument, the adaptation method causes a slightly higher NoC, but still lowers the failure rate. 
We also want to assure that this decreased performance does not stem from potential low-quality masks in SA-1B. For this purpose, we also tested HQ-SAM \citep{sam_hq} on our datasets, which is a slightly altered version of SAM that has been fine-tuned on high-quality human-annotated masks. In \Cref{tab:obscure85,tab:medical85,tab:obscure90,tab:medical90} we see that HQ-SAM performs drastically worse than SAM. We assume this to be the case due to a decrease in diversity which occurred during fine tuning. The novel segmentation head and HQ token have only ever been trained on the vastly smaller HQSeg-44K, rendering them particularly inept for the usage on unknown domains.
\section{Conclusion}
In our paper we applied the Segment Anything Model to uncommon situations. We did so for the specific task of interactive segmentation and evaluated appropriate metrics: The Number of Clicks (NoC) and the Failure Rate (FR). 
Despite the model being trained on the largest dataset for instance masks to date, we see considerable problems when confronting the model with data that differs from regular consumer images. In some situations the model failed to segment more than half of the objects in the dataset, as reflected by the Failure Rate. This inability to segment certain objects poses a crucial limit to the model.
In order to alleviate this problem we propose an efficient test time adaptation method. All techniques are restricted to using information that occurs during usage and do not require any previous fine-tuning on existing datasets. In addition to that, they only incur a minimal computational overhead in order to not hamper any potentially required real-time capabilities. 
With the help of our method we manage to lower the Failure Rate on twelve different datasets and lower the NoC on ten of them. We thus conclude that the information available during test time provides a useful tool when applying a foundation model such as SAM to uncommon domains.

{
    \small
    \bibliographystyle{ieeenat_fullname}
    \bibliography{main}
}
 

\end{document}